# MDIW-13: a New Multi-Lingual and Multi-Script Database and Benchmark for Script Identification

Miguel A. Ferrer , Abhijit Das , Moises Diaz , Aythami Morales , Cristina Carmona-Duarte ,Umapada Pal

## Abstract

Script identification plays a vital role in applications that involve handwriting and document analysis within a multi-script and multi-lingual environment. Moreover, it exhibits a profound connection with human cognition. This paper provides a new database for benchmarking script identification algorithms, which contains both printed and handwritten documents collected from a wide variety of scripts, such as Arabic, Bengali (Bangla), Gujarati, Gurmukhi, Devanagari, Japanese, Kan- nada, Malayalam, Oriya, Roman, Tamil, Telugu, and Thai. The dataset consists of 1,135 documents scanned from local newspaper and handwritten letters as well as notes from different native writers. Further, these documents are segmented into lines and words, comprising a total of 13,979 and 86,655 lines and words, respectively, in the dataset. Easy-to-go benchmarks are proposed with handcrafted and deep learning methods. The benchmark includes results at the document, line, and word levels with printed and handwritten documents. Results of script identification independent of the document/line/word level and independent of the printed/handwritten letters are also given. The new multi-lingual database is expected to create new script identifiers, present various challenges, including identifying handwritten and printed samples and serve as a foundation for future research in script identification based on the reported results of the three benchmarks.

## Introduction

With the ever-increasing demand for the creation of a digital world, many Optical Character Recognition (OCR) algorithms have been developed over the years. A script can be defined as the graphic form of the writing system used to write a statement. The availability of large numbers of scripts makes the development of a universal OCR a challenging task. This is because the features needed for char- acter recognition are usually a function of structural script properties and the number of possible classes or characters. The extremely high number of available scripts makes the task quite daunting, and as a result, most OCR systems are script-dependent [1]. Script identification is the initial cognitive process that occurs when a human read printed or handwritten texts.

 Perceptual rules ensure that humans focus on borders and corners to ensure accurate identification. When it comes to automatic script identification, our aim is to utilize features rooted in cognitive principles to achieve optimal results. In this paper, we propose using texture-based features on black-and-white images as the first step.

 These features will emphasize borders and corners, aligning with cognitive principles. The extracted features will then be inputted into machine learning schemes for script identification. In the second step, we will leverage deep learning classifiers that emulate the interconnected nature of human cognitive processes to perform the same task.

 Our approach involves comparing texture features and machine learning schemes with deep learning paradigms to establish a benchmark for the shared new multi-lingual MDIW-13 script identification database. This benchmark is expected to serve as a valuable resource for evaluating and comparing diverse script identification methods.

 The approach for handling documents in a multi-lingual and multi-script environment is divided into two steps: first, the script of the document, block, line, or word is estimated, and secondly, the appropriate OCR is used. This approach requires a script identifier and a bank of OCRs, at a rate of one OCR per possible script.

 Many script identification algorithms have been proposed in the literature. A survey published in 2010 with a taxonomy of script identification systems can be found in [2]. A more recent global study on state-of-the-art script identification can be found in [3]. Instead, the survey in [4] is focused on Indic Scripts. These surveys report novel performances of script identification methods based on pattern recognition strategies.

 Script identification can be conducted either offline, from scanned documents, or online if the writing sequence is available. Identification can also be classified either as *printed* or *handwritten*, with the latter being the more challenging. Script identification can be performed at different levels: page or document, paragraph, block, line, word, and character. An example for Indic scripts is given in [5].

As it is similar to any classical classification problem, the script identification problem is a function of the number of possible classes or scripts to be detected. Furthermore, any similarity in the structure of scripts represents an added challenge. If two or more scripts are very similar, then the identification complexity increases. For example, the Kan- nada and Telugu scripts are very similar and thus, lend them- selves to confusion in many cases. Although documents with two scripts represent the most common problem, documents with three and more scripts can also be found [6].

A unified approach based on local patterns analysis was proposed in [7] for script identification at line level and improved in [8] for word level. It was applied to video frames in [9]. In these cases, histograms of local pat- terns are used as features describing both the direction distribution and global appearance of strokes. In a further step, Neural Networks have demonstrated their capacity to extract highly discriminant features from images when enough data is available. Consequently, Neural Networks with Deep Learning have been explored in many tasks that involve document analysis. Specifically, in [10], the authors proposed a Discriminative Convolutional Neural Network (DCNN). Their approach combines deep features obtained from three convolutional layers. Their results, which registered performance gains of over 90% in a data- base with 13 scripts, demonstrate the feature extraction capacity of DCNN for script identification tasks.

Other approaches have explored similar or optimized architectures like Discriminative CNN [10]. An example is given in [11], where the authors stated that addressing the script identification problem with state-of-the-art Convolutional Neural Network (CNN) classifiers is not straight- forward, as they fail to address some key characteristics of scripts, e.g., their extremely variable aspect ratio. Instead of resizing input images to a fixed aspect ratio, the authors of [11] proposed a patch-based classification framework to preserve discriminative parts of the image.

To this end, they used ensembles of conjoined networks to jointly learn discriminative stroke-part representations and their relative importance in a patch-based classification scheme.

CNNs have further been applied to handwritten script recognition, as proposed in [12]. In that work, an architec- ture composed of two convolutional layers was employed. The results in a database containing 5 scripts demonstrate the potential of CNNs in either handwritten or printed text. Recurrent Neural Networks (e.g., Long Short-Term Memory Networks) have been explored in the context of Arabic [13] and Indic [14] script identification. These network architectures allow capturing sequential information and achieving state-of-the-art performance. Also, a combination of individually trainable small CNNs with modifications in their architectures was used in [15] for multi-script identification.

Further, the authors in [16] introduced the extreme learn- ing machine (ELM) technique, which generalizes the performance of neural networks. The authors studied this technique on 11 official Indic scripts and observed significant results when the sigmoidal activation function was used.

The power of CNN was also evidenced in [12] to identify Chinese, English, Japanese, Korean, or Russian scripts. The authors also evaluated whether the texts were handwritten or machine-printed and obtained excellent performances.

In summary, while most works claim identification rates exceeding 92%, each work, however, uses different datasets with different script combinations. Therefore, it is difficult to carry out a fair comparison of these different approaches. Moreover, the databases employed in related studies usually include two to four scripts. A few actually include an even higher number of scripts. The most popular scripts are Latin, Indian, Japanese, and Chinese, with Greek, Russian and Hebrew also featuring here and there [2]. A common database allowing a fair comparison of different algorithms would thus be desirable.

While building a dataset used to be a costly endeavor, it has become much simpler and easier today, even though the task remains arduous and laborious. For instance, documents from different scripts can be generated using the Google Translate application, as in [8], for example. However, in this case, the font, size, and background of the generated document will be the same, which is unrealistic.

To alleviate this drawback, this paper aims to offer a database for script identification, which consists of a wide variety of some of the most commonly used scripts, collected from real-life printed and handwritten docu- ments. Further, along with the database, its benchmark- ing with texture-based features and deep learning are also showcased.

The printed documents in the database were obtained from local newspaper and magazines, and there- fore, comprise different fonts and sizes and cursive and bold text. A sample of the newspaper used can be seen in Fig 1. The handwritten part was obtained from volunteers from all over the world, who scanned and shared their manuscripts. A few samples of the handwritten documents can be seen in Fig 2.

The following three benchmarks of this database are provided for script identification using different hand-crafted features: Local Binary Pattern [17], Quad-Tree Histogram of Templates [18], and Dense Multi-Block Local Binary templates with a Support Vector Machine as a classifier [19]. These script identifiers were used in a document analysis context in [4] and [5]. A benchmark with Deep Learning techniques is also included in our study to demonstrate the usefulness of this database to train deep models.

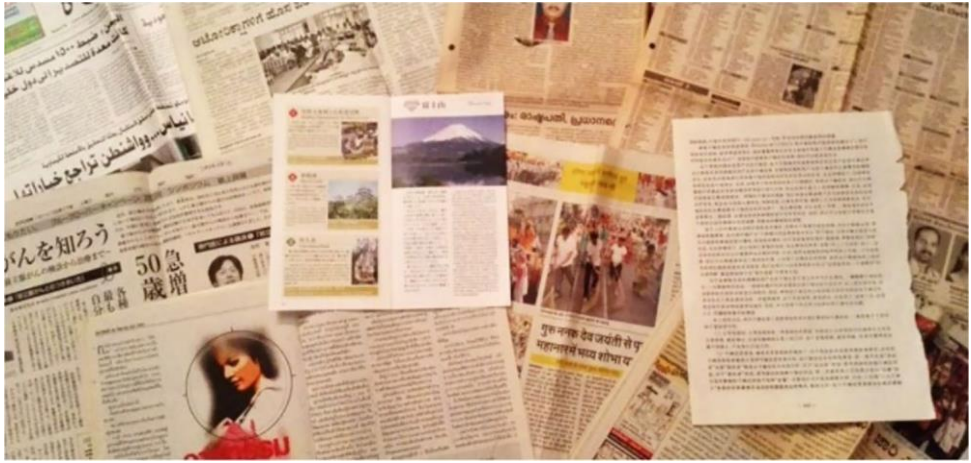

**Fig. 1** *Samples of newspaper used for the dataset.*

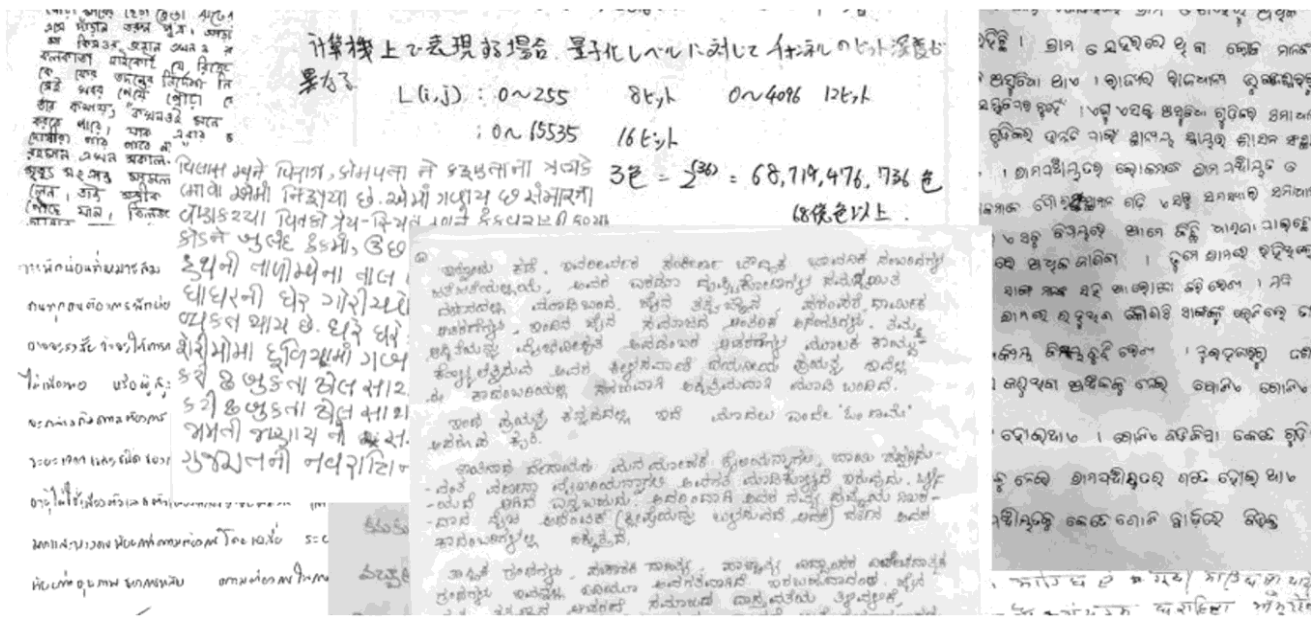

**Fig. 2** *Samples of handwritten documents used for the dataset.*

As a summary, the contributions of the work are listed as follows:

1. A freely accessible multi-lingual database towards script identification called MDIW-13 (Multi-lingual and Multi-script Document Identification in the Wild. Num- ber 13 refers to the number of scripts in the dataset).
2. The database provides the possibility of handwritten and printed script identification.
3. The database allows script identification at document, line, and word levels.
4. The database enables cross-training, e.g., train with printed and test with handwriting; train with lines and test with words, among others.
5. A benchmark with different standard parameters and classifiers is given for the sake of comparison.

## Previous Works on Public Databases

The research community is interested in script identification as it can help in different document analysis tasks, such as OCR, handwriting recognition, document analysis or writer identification [20]. However, the number of script identification databases available is limited, so there is a significant need for publicly available databases. Regarding the number of scripts, size, and availability of datasets for script identification, the most popular pub- lic databases contain only Roman and Arabic scripts. An example includes the database of the Maurdor project [21], which is contemporary to the MALIS-MSHD [22]. Other ones can be also used for script identification although they are devised for writer recognition [23]. Also exist databases of printed script [24]. Roman, Bengali and Devanagari databases were compiled in [25]. The authors proposed bi- script and tri-script word-level script identification bench- marks studying the performances in several classifiers. The literature also considers databases with peculiar scripts, which have not been thoroughly investigated in handwrit- ing. An example can be seen in [26], where an Indic data- base includes the Meitei Mayek script. The SIW-13 [27] is a script identification benchmark with ten different scripts composed of printed text obtained from natural scene images. SIW-13 consists of 10 scripts, including English, Greek, Hebrew, Russian, Arabic, Thai, Tibetan, Korean, Kannada, Cambodian, Chinese, Mongolian, and Japanese. Also, in [28] is found PHDIndic_11, a publicly available dataset focused on 11 official Indic Scripts, which are used in the 22 official languages in India. Previous existing data- bases are summarized in Table 1.

*Table 1* *Summary of public script identification databases (H &P = Handwritten and Printed samples)*

| Ref. | H &P | #scripts | #language | #words | #docs |
|------|------|----------|-----------|--------|-------|
| [21] | Yes | 2 | 3 | - | 2.5K |
| [22] | No | 2 | 2 | - | 1.2K |
| [23] | No | 2 | 2 | - | 1K |
| [24] | Yes | 2 | 2 | - | 5K |
| SIW[27] | No | 13 | 13 | 13K | 7.7K |
| [28] | No | 11 | 22 | 19K | 1.5K |
| [29] | No | 3 | 4 | 5.6K | - |
| [30] | No | 4 | 4 | 104K* | 0.7K |
| Our MDIW | Yes | 13 | 13 | 87K | 1K |

H &P: Handwritten and Printed documents

*word/subword

The new database built in this work, MDIW-13, rep- resents a step forward in the field of script identification, with 13 scripts and over 87,000 handwritten and printed words. The main difference between our work and existing databases lies in a large number of scripts employed in the proposed dataset. Some of these 13 scripts are pretty similar, whereas others are somewhat different. Also, some of them can be found in real applications in countries like India, where many Indian and even non-Indian scripts can be found in border control, access, courier companies, or document analysis. This property makes the MDIW-13 database more versatile and interesting in Indic environments. Furthermore, MDIW-13 is composed of text extracted from documents, which is carefully preprocessed to eliminate covariates from background and acquisition protocols.

Another contribution of this paper is to provide a bench- mark with well-known and easy to replicate script identifiers. In this case, the benchmark leads to studying the performance impact when the training set uses words or lines or pages or a combination of all three. This kind of experiment is a possibility offered by MDIW-13.

The rest of this paper is organized as follows: Sect. 2 describes the database and its different features. While Sect. 3 describes the proposed script identifiers for bench- marking purposes, Sect. 4 gives the benchmarking design and experiment results. Section 6 and 7 close the paper with a discussion and a conclusion.

## MDIW-13: A New Database for Script Identification

The proposed database consists of printed and handwritten samples from a total of 113 documents, which were scanned from local newspaper and handwritten letters and notes. From these documents, a total of 13,979 lines and 86,655 words from 13 different scripts were extracted. The database is offered with the raw data from direct digitalization and after the preprocessing carried out here. This database can be freely downloaded for research purposes.

### Main Challenges in Data Collection

Probably, the main challenge in this work was to obtain the data, especially from newspaper, because of the wide variety of scripts involved.

It is possible that some documents for each script could contain some sort of watermark owing to the fact that each document came from a different original native location. This poses a risk of the document watermark, rather than the script, being recognized, which could be the case with a deep learning-based classifier.

Segmenting text from the backgrounds of some documents was challenging. Even with state-of-the-art segmentation techniques used, the result was unsatisfactory and included a lot of salt and pepper noise or black patches, or some parts of the text were missing.

To avoid these drawbacks and provide a dataset for script recognition, all the documents were preprocessed and given a white background, while the foreground text ink was equalized. Furthermore, all documents were manually examined. Both original and processed documents are included in the database.

To conduct experiments on script recognition at different levels (i.e., document, line and word), each document was divided into lines and each line into words. In this division, a line is defined as an image with two or more words, and a word is defined as an image with two or more char- acters. It is worth highlighting that the whitespaces were unaltered in any case since the importance of their use in script identification.

In the following subsections, specific challenges in digitalizing both printed and handwritten documents are highlighted.

### Main Challenges in Digitizing Printed Documents

The part of the database from printed documents was acquired from a wide range of local newspaper and maga- zines to ensure that the samples would be as realistic as   possible. The newspaper samples were collected mainly from India (as a wide variety of scripts are used there), Thailand, Japan, the United Arab Emirates, and Europe. A few examples of the printed documents used are shown in Fig 1. The database includes 13 different scripts: Arabic, Bengali, Gujarati, Gurmukhi, Devanagari, Japanese, Kan- nada, Malayalam, Oriya, Roman, Tamil, Telugu, and Thai. The newspaper were scanned at a 300 dpi resolution. Paragraphs with only one script were selected for the database (paragraph here means the headline and body text).

These paragraphs included multiple fonts, letter styles with italics or bold formats. Nevertheless, some newspaper mix different scripts in the same text. For instance, an Arabic number or a Latin character could be found in a Devanagari script. In these cases, it was tried not to mix those scripts in a single part of the database.

Further, it was tried to ensure that all the text lines were not skewed horizontally. All images were saved in *png* format, using the *script_xxx.png* naming convention, with script being an abbreviation or memo for each script, and *xxx*, the file number starting at 001 for each script. The scripts, abbreviations, and the number of documents for each script are illustrated in Table 2. Further information about the dataset can be found in Tables 10, 11 and 12 in the Annexes.

*Table 2*  *Database figures*

| Script | Abbrev | Handwritten | | | Printed | | |
|---|---|---|---|---|---|---|---|
| | | Docs | Lines | Words | Docs | Lines | Words |
| Arabic/Per | *Arab* | 48 | 621 | 3940 | 51 | 1082 | 6202 |
| Bengali | *Ban* | 67 | 1486 | 9320 | 51 | 466 | 2557 |
| Gujarati | *Guj* | 3 | 41 | 181 | 32 | 384 | 2211 |
| Gurmukhi/Punjabi | *Gurm* | 6 | 111 | 700 | 115 | 1062 | 9104 |
| Devanagari | *Hind* | 21 | 230 | 1457 | 47 | 397 | 2782 |
| Japanese | *Jap* | 20 | 121 | 441 | 80 | 559 | 1814 |
| Kannada | *Kan* | 15 | 377 | 1995 | 53 | 582 | 2157 |
| Malayalam | *Mal* | 12 | 211 | 719 | 70 | 706 | 4320 |
| Oriya | *Ori* | 50 | 1136 | 7847 | 42 | 548 | 2309 |
| Roman | *Rom* | 90 | 750 | 4308 | 56 | 961 | 7627 |
| Tamil | *Tam* | 14 | 276 | 1430 | 46 | 301 | 2118 |
| Telugu | *Tel* | 10 | 154 | 801 | 49 | 483 | 2126 |
| Thai | *Tha* | 26 | 473 | 4472 | 61 | 461 | 3717 |
| | Total: | 382 | 5987 | 37611 | 753 | 7992 | 49,044 |

Docs: number of documents; Lines: number of lines; Words: number of words

*Main Challenges in Digitizing Handwritten Documents*

Similarly to the printed part, the handwritten database also included 13 different scripts: Persian as Arabic, Bengali, Gujarati, Punjabi, Gurmukhi, Devanagari, Japanese, Kan- nada, Malayalam, Oriya, Roman, Tamil, Telugu and Thai.

To collect them, several invitations were sent to several native researchers and colleagues from different countries, who were capable of writing documents in their respective scripts, asking for handwritten letters. Each volunteer wrote a document with their pen and with no restrictions on the paper type used. Next, they digitized these documents on unspecified devices and without the limitation of scanning settings, such as resolution, and then sent them to us by e-mail. Consequently, the documents had large ink, sheet and scanner quality variations. All these uncontrolled conditions meant constructing a database as close to the wild as possible. Note that the Roman sheets came from the IAM handwritten database [31]. Some examples are shown in Fig 2.

*Background and Ink Equalization*

Due to the broad quality range of the documents, a two- step preprocessing was performed. In the first step, images are binarized by transforming the background into white, while in the second step, an ink intensity equalization is performed.

Because the background texture, noise, and illumination conditions are primary factors that degrade document image binarization performance, an iterative refinement algorithm was used to binarize [32]. Specifically, the input image is initially transformed into a Bhattacharyya similarity matrix with a Gaussian kernel, which is subsequently converted into a binary image using a maxi- mum entropy classifier. Then a run-length histogram is used to estimate the character stroke width. After noise elimination, the output image is used for the next round of refinement, and the process terminates when the estimated stroke width is stable. However, some documents are not correctly binarized, and in such cases, a manual binarization is performed using local thresholds. All the documents were reviewed, and some noise was removed manually.

Sometimes, collaborators made mistakes during the writing of the letter. Such mistakes resulted in blurred handwriting with scribbles in some parts of the letters which were identified and repaired by adding white boxes to these scribbled parts of the documents.

For ink equalization, an ink deposition model proposed in [33] was used. All the black pixels on the binarized images were considered ink spots and correlated with a Gaussian width of 0.2 mm. Finally, the image was equalized to duplicate fluid ink, as in [34]. The result can be seen in Fig 3.

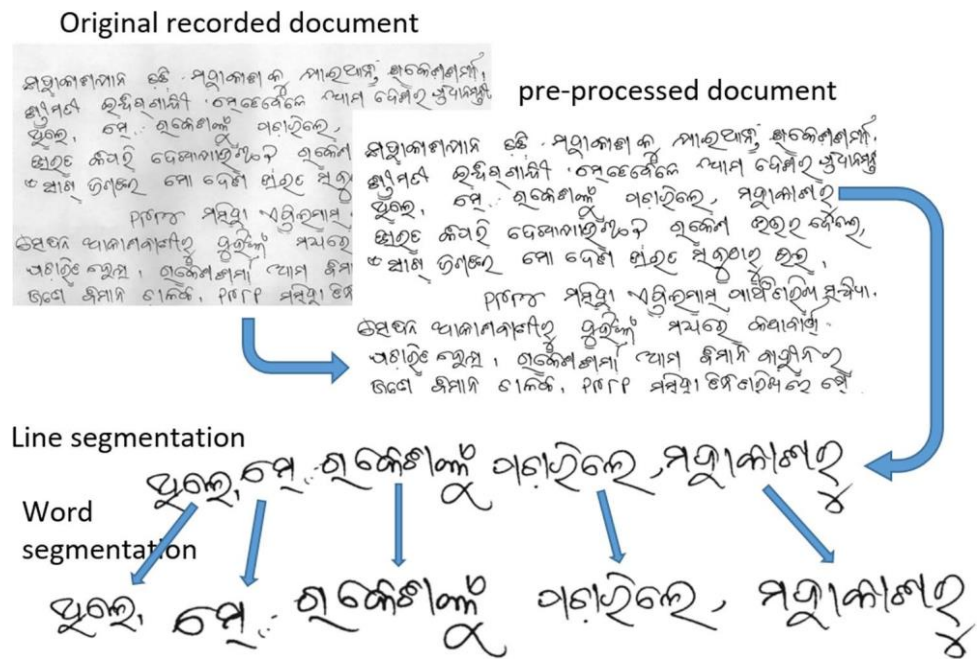

*Fig. 3  Pre-processed database and line and word segmentation*

### Text line Segmentation

For the lines from a document to be segmented, they must be horizontal; otherwise, a skew correction algorithm must be used [17].

For line segmentation, each connected object/component of the image is detected, and its convex hull obtained, as shown in Fig 4. The result is dilated horizontally in order to connect the objects belonging to the same line (see Fig. 4) and each connected object is labelled. The next step is a line-by-line extraction, performed as follows:

1. Select the top object of the dilated lines and determine its horizontal histogram.
2. If its histogram has a single maximum, then it should be a single line, and the object is used as a mask to segment the line (see Fig 4).
3. If the object has several peaks, it is assumed that there are several lines. To separate them, the following steps are followed:
4. The object is horizontally eroded until the top object contains a single peak.
5. The new top object is dilated to recover the original shape and is used as a mask to segment the top line.

6.The top line is deleted, and the process is repeated from step 1 to the end.

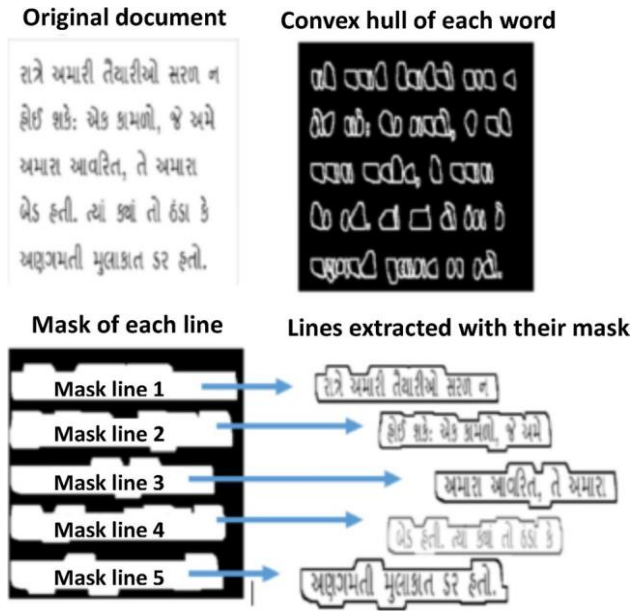

**Fig. 4** *Line detection procedure*

This automatic segmentation procedure was initially used. Later, each line was manually examined. Any lines that had been wrongly segmented were manually repaired. The lines were saved as image files and named using the *script_xxx_yyy.png* format, where *yyy* is the document number and script is the abbreviation for the script, as previously mentioned. Figure 3 presents an example of a segmented line for handwriting. These images are saved in grayscale format. The number of lines per script can be seen in Table 2.

### *Word Segmentation*

The words were segmented from the lines in two steps, with the first step being completely automatic. Each line was converted to a black and white component, a vertical histogram was obtained, and points where the value of the histogram was found to be zero were identified as the gaps or the intersection. Gaps wider than one-third of the line height were labelled as word separations.

In the second step, failed word segmentations were manually corrected. Each word was saved individually as a black and white image. The files were named using the *script_xxx_yyy_zzz.png* format, with *zzz* being the word number of the line *script_xxx_yyy*. For instance, a file named *roma_004_012_004.png* contains the black and white image of the fourth word on the 12th line of the 4th document in Roman script. An example of the segmentation result can be seen in Fig. 3. The number of words per script is shown in Table 2.

In Thai and Japanese, word segmentation is conducted heuristically because their lines consist of two or three long sequences of characters separated by a greater space. This is because there is generally no gap between two words in these scripts, and contextual meaning is generally used to decide which characters comprise a word. Since we did not conduct text recognition and no contextual meaning is applied in the current database, the following approach for pseudo-segmentation of Thai and Japanese scripts was used after sought advice from native Thai and Japanese writers: for each sequence of characters, the first two characters are the first pseudo-word; the third to the fifth characters are the second pseudo-word; the sixth to the ninth characters are the third pseudo-word, and so on, up to the end of the sequence.

It should be noted that in this work, our intention is not to develop a new line/word segmentation system. Only a simple procedure is used to segment lines and words in a bid to build our database. In this way, a semi-automatic approach is worked out, with human verification and correction in the case of erroneous segmentation.

**Script Identifiers**

For database benchmarking, an automatic script identifier is required. For a more general benchmarking of the database, up to four automatic script identifiers are used. The two firsts are based on the classical feature-classifier structure, and the last two are based on deep learning. Our motivation in defining the benchmarks is that they are easy to replicate by third parties, allowing them to establish a baseline in these three cases. To this aim, the systems are accessible in several toolboxes under different programming languages. In feature-classifier script identifiers, the script feature extractors used in this section are based on local patterns. Specifically, we used Local Binary Patterns (LBP) [17], Quad-Tree histograms [18], and Dense Multi-Block LBPs [19]. Such features can be seen as constituting a unifying approach, thus bringing together the traditional appearance and structural approaches.

When these techniques are applied to black and white images, local patterns can be considered as the concatenation of the binary gradient directions. The histogram of these patterns contains information on the distribution of the edges, spots, and other local shapes in the script image, which can be used as features for script detection. The following section describes the features used for script identification. The classifier used for script identification, which is a Support Vector Machine (SVM) [35], is also described. For script identification based on deep learning, two popular state-of-the-art image recognition architectures based on Convolutional and Residual layers are used for benchmarking.

*Local Binary Patterns for Script Detection*

*Local Binary Patterns*: The original LBP [17] operator labels the pixel of an image by thresholding the $3 \times 3$ neighborhood around each pixel and concatenating the results binomially to form a number. Assume that a given image is defined as I(Z)=I(x,y) . The LBP operator transforms the input image to LBP(Z) as follows:

$$LBP(Z_C) = \sum_{p=0}^{7} s\left(I(Z_p) - I(Z_C)\right) 2^p \qquad (1)$$

$where\ s(l) = \begin{cases} 1, & l \geq 0 \\ 0, & l < 0 \end{cases}$ is the unit step function and I($Z_p$) is the 8-neighborhood around I(Zc) , and p represents the order of the considered neighbor. In this paper, we set p to 3, corresponding to an 8-neighbor configuration.

LBP feature: The LBP(Z) code matrix contains information about the structure to which the pixel belongs –a stroke edge, a stroke corner, a stroke end and so on. It is assumed that the distribution of these structures defines the script. The distribution is obtained as the histogram of the LBP(Z), named $h_{LBP}$ . As the histogram is a function of the size of the image, it is normalized as $hn_{LBP} = h_{LBP} / \sum h_{LBP}$ . The length of this vector is 255 since the LBP value for the background is discarded.

The problem with the histogram is that it leads to a loss of spatial distribution of the structures. To include the spatial distribution in the LBP feature, the image is divided into a number of zones so as to calculate the histogram in each zone as a vector $hn_{LBP}$ , and then concatenating them. After several experiments were conducted, and a range of smaller and larger zone sizes were tested, the best performance was obtained when dividing the lines and words into three equal horizontal regions, which overlapped by 30%. Thus, the vector $H_{LBP} = [hn^1_{LBP}, hn^2_{LBP}, hn^3_{LBP}]$ of 765 components was worked out.

Finally, this vector of size 765 was reduced to 255, start counting from the zeroth component, by calculating the DCT of $H_{LBP}$ and by selecting from the second to the 256th component. This new vector is the LBP feature used to identify scripts in the cases of lines and words. An example of this procedure is illustrated in Fig 5.

In the case of a full document with several lines, the LBP features of all the lines were combined at the score level.

*Quad-Tree Histogram of Templates for Script Detection*

In this section, it is proposed a new and efficient feature for script identification. It is based on a quad-tree computation of the Histogram of Templates (HOT). It was introduced for signature verification in [18]. Specifically, this feature is an extension of the HOT, which is introduced to highlight local directions.

The implementation of the HOT employs a set of 20 templates to describe the segment orientations by comparing the positional relationship between a pixel and its neighborhood references. Specifically, a sliding window covering $3 \times 3$ pixels is applied to the text image to count the number of pixels that fit this template. The resulting counts constitute the histogram of the templates. In [18], HOT is computed by considering the pixel and gradient information. This vector is calculated in the following steps:

1. Pixel information-based HOT (P-HOT). There are 20 possible templates, and each template corresponds to a possible combination of adjacent pixels $Z_1$ and $Z_2$ with pixel $Z =(x,y)$ . For each template and pixel $Z$, if the grey value $I(Z)$ is greater than the grey value of the two adjacent pixels $I(Z_1)$ and $I(Z_2)$ , then add 1 to the value of this template. In other words, the following condition should be satisfied:

$$I(Z) > I(Z_1) \wedge I(Z) > I(Z_2). \tag{2}$$

The vector of the tally of these 20 templates is termed a histogram of templates, which is the feature vector.

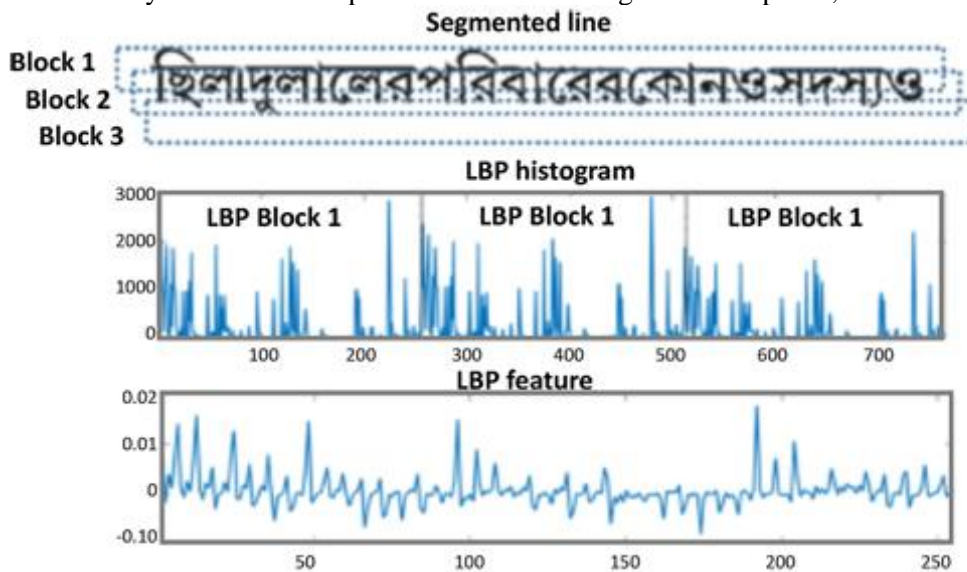

*Fig. 5 LBP features for script identification*

2. Gradient information-based HOT (G-HOT): For each template, if the gradient magnitude $Mag(Z)$ of a pixel $Z =(x, y)$ is greater than the gradient magnitude of the two adjacent, pixels, i.e. $Z$, matches the following condition:

$$Mag(Z) > Mag(Z_1) \wedge Mag(Z) > Mag(Z_2) \tag{3}$$

In this case, we then add 1 to the value of this template. There are 20 possible combinations of adjacent pixels $Z_1$ and $Z_2$ for each pixel $Z$, and so there are 20 templates. Similar to P-HOT, the vector of the tally of the 20 templates is known as the gradient histogram of templates, which acts as the feature vector.

The HOT template consists of the 20 values of the P-HOT feature concatenated with the 20 values of the G-HOT feature, for a total of 40 values. To facilitate the verification process, after the HOT calculation for each region, it is per formed an L2 normalization on the 40 values of the HOT to scale in the margin between 0 and 1.

The quad-tree structure considers the spatial property of a local shape by dividing it into four cells at different levels. The center of gravity of the pixels is assigned to the center of the equi-mass partition. This overcomes empty computing cells, especially at deeper levels. Therefore, HOT is locally computed at each level of the quad-tree structure, while the whole image feature is obtained by concatenating all local HOT features.

Heuristically, it was used the HOT features at the first and second quad-tree levels. There is the full image at the first level, while at the second, there are four partitions. Hence, there are 5 HOT features, which run to a 200-dimensional feature vector. An example of this procedure is shown in Fig.6.

### Script Features Based on Dense Multi-Block LBP Features

Dense Multi-Block LBPs (D-LBP) are new features that have recently been proposed for script identification, and they are derived from LBP, as indicated by Equation 1. They are based on a spatial pyramidal architecture of the multi-block LBP (MBLBP) histograms proposed in [36]. We chose this classifier for its performance properties, making it suitable for our benchmark. It is well-suited for cognitive computation, approximating the human cognitive process of information selection. Additionally, our article includes other novel classifiers, allowing us to showcase a wide performance spectrum and analyze our database thoroughly.

Specifically, an image I of $n_x$ rows and $n_y$ columns, at level $l = 1, 2,...,L$, is divided into $N_x$ l by $N_y$ l patches of height hl and width $w_l$. The patches are uniformly distributed in the image. For each patch, the histogram of MBLBP descriptors at different scales is worked out. The feature consists of all the concatenated histograms, which result in a feature of dimension $\sum_{l=0}^{L} 256 \, s \, N_l^x \, N_l^y$.

In our case, for script identification, it is heuristically defined L = 2 and s = 4. At the first level, $N_1^x = 1$, $N_1^y$, $h_1 = n_x$ and $w_1 = n_y$ at the second level, $N_2^x = 3$, $N_2^y = 3$, $h_2 = 0.5 \, n_x$ and $w_2 = 0.5 \, n_y$, and so the 9 ($3 \times 3$) patches are 25% overlapped. Hence, the final feature vector dimension is 10,240. An example of the distribution of the patches is shown in Fig. 7 for a Gurumukhi word. This feature vector was implemented using the Scenes/Objects classification toolbox freely available in the Matlab central files exchange.

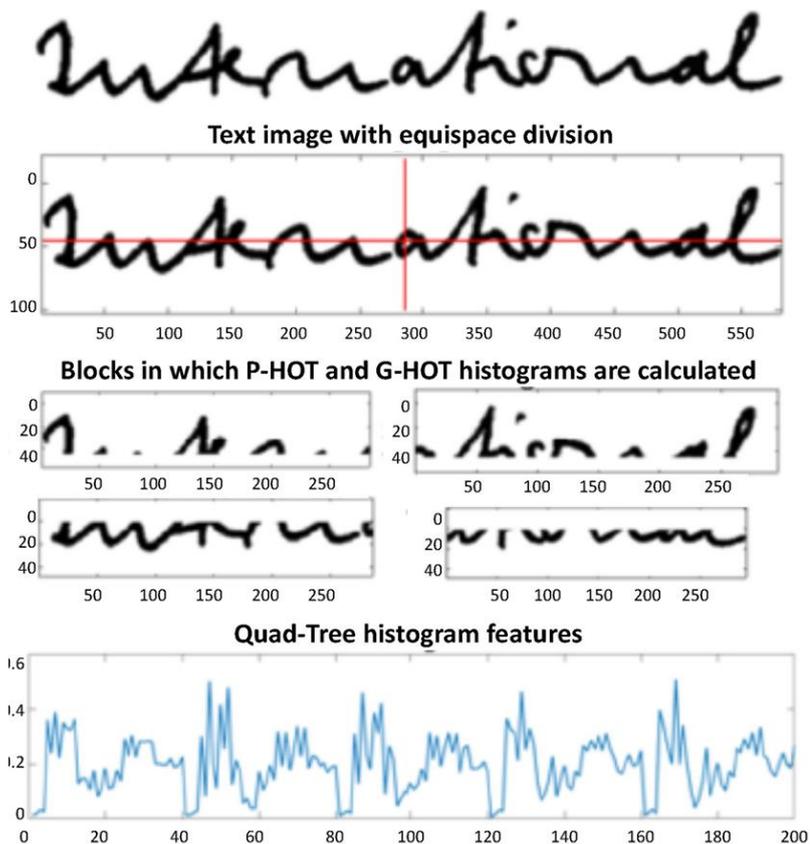

*Fig. 6 Quad-Tree Histogram for script identification*

*Classifier*

A Support Vector Machine (SVM) was used as a classifier because of the large dimension of the feature vectors. An SVM is a popular supervised machine learning technique that performs an implicit mapping into a higher dimensional feature space. This is the so-called kernel trick. After the mapping is completed, the SVM finds a linear separating hyperplane with maximal margin to separate data from this higher dimensional space. Least Squares Support Vector Machines (LS-SVM) are reformulations of standard SVMs which solve the indefinite linear systems generated within the latter. Robustness, sparseness, and weightings can be imposed on LS-SVMs where needed, and a Bayesian framework with three levels of inference is then applied [35]. Although new kernel functions are being proposed, the most frequently used kernel functions are the linear, polynomial, and Radial Basis Function (RBF). The present study uses the RBF kernel for LBP and Quad-Tree features and a linear kernel for Dense LBP. SVM or LS-SVM makes a binary decision, while, in this study, multi-class classification for script identification is carried out by adopting one-against-all techniques. Grid searches were carried out on the hyper-parameters in 2-fold cross-validation to select the parameters in the training sequence.

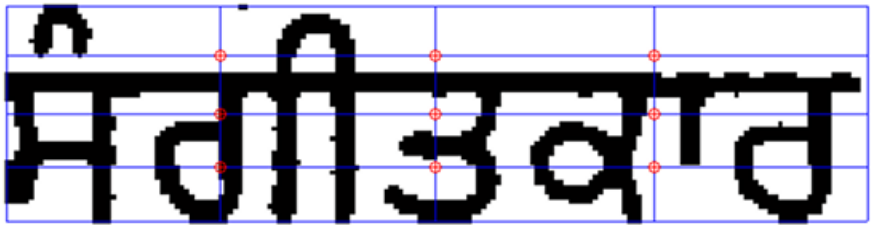

*Fig. 7 Example of the 12 overlapped patches on a word; Red circles: patch centers*

**Deep Neural Network Architectures**

Deep Neural Networks have demonstrated their potential in many computer vision tasks when sufficient data is available. They are proposed to evaluate the usefulness of this new database in training deep learning architectures. The more than 30K labelled words included in this database constitute a valuable new resource for the scientific community.

Our experiments employed two popular state-of-the- art Convolutional Neural Networks architectures based on VGG [37] and ResNet [38] models. These architectures have been chosen as examples of data-driven learning models employed in image classification challenges. During the last decade, Deep Convolution Neural Networks has boosted the performance of Computer Vision applications, including text classification [39, 40]. The VGG architecture used in our experiments is based on the traditional 2D convolutional layers. The ResNet model improves the traditional convolutional architectures introducing the residual connections between convolutional layers (i.e., shortcuts between layers). The residual connections improve the training process of the network, providing higher performance. The visual information of the strokes such as directionality, curvature, frequency, or density is critical to classify the different scripts. In both cases (VGG and ResNet), the 2D convolutional filters learned during the trained process present a great capacity to model such visual patterns.

Each input image is subsampled at the preprocessing step into 60x 60-pixel sub-images using a sliding window (50% overlap). In order to improve the generalization capability of the model, data augmentation techniques are applied (shear, zoom, width, and height shift).

The first architecture evaluated is a VGG architecture. This network is composed of two convolutional layers followed by one fully connected layer with dropout (0.25) and 13 units (softmax activation). The ReLU (Rectified Linear Unit) activation function was used in all hidden layers and a max-pooling layer with a filter size of 2 x 2 after each convolutional layer. The first convolutional layers have 32 filters of size 3x 3 and stride 1, and the second convolutional layer has 64 filters of 3 x 3. This network comprises more than 3 M parameters.

The second architecture is a Residual Neural Network architecture. This network comprises three convolutional blocks and a dense output layer (13 output units and soft- max activation). The first convolutional block is composed of a convolutional layer (64 filters of size 7 x 7), and a 3 x 3 max pool layer (stride 2). The second and third blocks con- sist of identity and convolutional blocks. Our identity block includes a series of three convolutional layers with a bypass connection between the input of the identity block and the output of the third convolutional layer. The second convolutional block includes three convolutional layers (64, 64, and 256 filters of size 1 x 1, 3 x 3, and 1 1 respectively), a convolutional layer shortcut (128 filters of size 1 1), two identity blocks (64, 64, and 256 filters of size 1 x 1, 3 x 3, and 1 1 respectively) without the bypass connection. The third block has three convolutional layers (128, 128, and 512 filters of size 1 x 1, 3 x 3, and 1 1 respectively); and three identity blocks with this same series of filters per convolution. Batch normalization and a ReLU activation function after each convolutional layer were employed. This network comprises more than 1.5M parameters.

The following are the implementation details of the train- ing for both architectures: batch size of 128, Adam optimizer with a 0.001 learning rate, random initialization of the weights, and a number of epochs equal to 30 and 10  for handwritten and printed samples, respectively (printed models converge faster than handwritten ones). Both VGG and ResNet models were trained from scratch (i.e. we have not used pre-trained models). The architectures (i.e., number of layers, number of neurons per layer, activation functions) and the hyperparameters (i.e., optimizer, batch size, epochs, etc...) presented in this work are the results of several experiments. We have prioritized the configuration with the best performance and a lower number of parameters (i.e., fewer layers and neurons) during the experimentation. Further- more, we have discarded the use of pre-trained models to guarantee a fair comparison between benchmarks (i.e., the same data was used to train all three benchmarks).

### Benchmarking: Experiments

The benchmarking consists of classification experiments with the above-described techniques to estimate the script of a given document or line or a word among those included in the dataset. It should be borne in mind that the present benchmark attempts to measure the reach and range of the database built with well-known state-of-the-art classifiers and that it is not aimed to propose a new script identifier.

Three different benchmarks were constructed for this estimation. The first one uses a classifier based on a score level combination of LBP and Quad-Tree features. The second one is based on Dense Multi-Block LBP features. It is worth pointing out that the combination of these two systems improves the performance by about 10%. Finally, the third is constructed with two popular Deep Neural Network architectures (DNN). The three benchmarks are illustrated in Fig. 8, where we utilized LS-SVM for both tasks and combine them at the score level.

### Training Sequences

Defining the training sequences is paramount for a fair comparison of results. Thus, the classifiers for each printed and handwritten document script should be trained as similarly as possible. However, a database of handwritten or printed documents is inherently unbalanced because each of its constituent documents contains a different number of lines, the lengths of the lines are different, and the word sizes differ between the scripts. Therefore, training each classifier with a similar number of documents, lines, or words does not guarantee equality of training or the fair comparison of results. Consequently, instead of training each classifier with a given number of documents, lines, or words, it was decided to train them with a similar number of pixels. In this way, one classifier was trained with 100 images and other with 150 because the training images of the second classifier contain less text than those of the first classifier.

The primary reason is that our approach required train ing all classifiers with an equal amount of information. To quantify this information, we conducted tests using various entropy measures, such as Shannon entropy, on a subset of the database. Interestingly, our analysis revealed that the outcomes in terms of selecting the number of images for training and testing per script were comparable to the pixel count. Counting the number of pixels proved to be a more efficient and practical approach. Consequently, we opted to employ the pixel count as a criterion for determining the appropriate number of images to train each script.

In analyzing the database, it was heuristically decided to train each classifier with several images whose accumulated number of pixels would be approximately 2 M. The numbers of documents used to train each classifier are shown in Table 4. This training sequence breaks down into the follow ing proportions: 21.03% of handwritten words, 21.82% of handwritten lines, and 15.06% of handwritten documents. In the printed dataset, it was assumed a training scenario with 51.06% of documents, 45.2% of lines, and 45.85% of printed words. Therefore, there is room for a statistically meaningful test. Further information about the training partition of the dataset can be found in Tables 13, 14, and 15 in the appendix of this article.

To ensure experimental repeatability, it was predeter mined training and test sequences. The training images appeared first and in numerical order (e.g., the first 18 Devanagari handwritten documents or the first 256 Arabic printed lines, or the first 1608 Bengali printed words, etc.), and the rest of the images were used for testing.

Therefore, there were the next six training sequences: printed documents, printed lines, printed words, handwrit ten documents, handwritten lines, and handwritten words. Similarly, there were the following six testing sequences: printed documents, printed lines, printed words, handwritten documents, handwritten lines, and handwritten words. The twelve sequences were disjointed. It should also be noted that all the experiments reported were separately tested with these testing sequences.

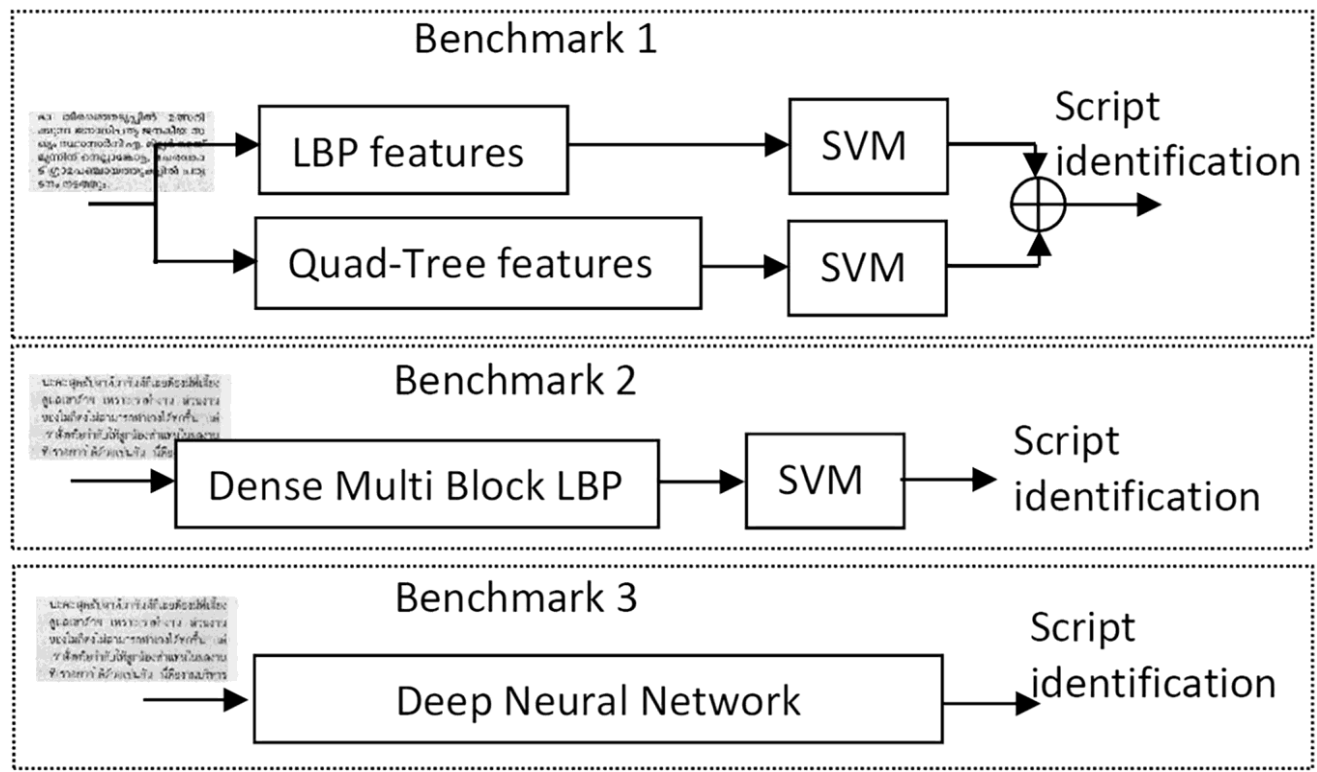

***Fig. 8*** *Benchmarks constructed in the paper*

### *Description of the Tasks*

Each benchmark was evaluated by performing three tasks, which depend on the different training sequences and are summarized in Table 3. The test sequence is the same for each task, which is composed of six different data types: printed documents, printed lines, printed words, handwritten documents, handwritten lines, and handwritten words. These were the remaining specimens of the database, which were not used in training, as shown in Table 4.

TASK 1: This task aims to study the behavior of the database at the document, line, and word levels [41] for printed and handwritten documents separately. Hence, each classifier is oriented to a specific type of document (document, line, or word and printed or handwritten) per script.

*Table 3 Description of the tasks per benchmark*

| | Training with: |
| --- | --- |
| | Handwritten docs |
| | Handwritten lines |
| Task 1 (one classifier per | Handwritten words |
| type of image and script) | Printed docs |
| | Printed lines |
| | Printed words |
| Task 2 (one classifier for hand-written and another for printed per script) | Handwritten docs, lines and words |
| | Printed docs, lines and words |
| Task 3 (a single | Printed and Handwritten |
| classifier per script) | docs, lines and words |

Evaluation protocol of task 1: It requires as many classifiers as scripts and type of image: document, line, and word in both printed or handwritten modality. Also, as the data base includes handwritten and printed specimens, the total number of classifiers used in this task is $13 \times 3 \times 2 = 78$ . These have been individually trained with the number of images indicated in Table 4. Once the remaining images are tested, a $13 \times 13$ confusion matrix is worked out with the performances given in percent (%) of each type of image per script. Then, the final identification performance is obtained as the average of the main diagonal of the performances in percentage.

TASK 2: This task aims to study the database behavior when the script classifier is oriented to being printed or handwritten, regardless of the type of document. Con sequently, the training of a particular classifier will include documents, lines, and words of a specific script and type of document: printed or handwritten.

Evaluation protocol of task 2: In this task, each classifier was trained with three training sequences of handwritten or printed documents for each script. In total, $13 \times 2 = 26$ classifiers were trained. It should be noted that the train ing words belong to the training lines, which in turn correspond to the training documents. The trained classifiers were tested with the six types of testing images, regardless of their type and modality. Then $13 \times 13$ confusion matrices were obtained in each case for each script, which were averaged in the same terms as in Task 1. Following the same strategy as task 1, the main diagonal values were averaged from the script confusion matrices to obtain the final performance.

TASK 3: The goal of this task is to study the database behavior independently of the input to the script classifier. This can be a printed or handwritten document or line or word.

Evaluation protocol of task 3: It requires 13 classifiers, one per script, which are trained with all types of documents, both printed and handwritten. After the testing, the final performance is obtained in the same terms as tasks 1 and 2.

*Table 4 Number of documents, lines and words for training*

| Script | Abbrev | Handwritten | | | Printed | | |
|---|---|---|---|---|---|---|---|
| | | Docs | Lines | Words | Docs | Lines | Words |
| Arabic/Per | *Arab* | 5 | 88 | 570 | 14 | 256 | 1996 |
| Bengali | *Ban* | 3 | 55 | 401 | 27 | 234 | 1608 |
| Gujarati | *Guj* | 2 | 32 | 144 | 22 | 190 | 1229 |
| Gurmukhi/Punjabi | *Gurm* | 4 | 88 | 560 | 39 | 468 | 3629 |
| Devanagari | *Hind* | 15 | 184 | 1165 | 33 | 215 | 1706 |
| Japanese | *Jap* | 4 | 96 | 352 | 64 | 447 | 1451 |
| Kannada | *Kan* | 3 | 122 | 872 | 38 | 302 | 1183 |
| Malayalam | *Mal* | 9 | 168 | 575 | 26 | 314 | 2370 |
| Oriya | *Ori* | 3 | 49 | 333 | 25 | 348 | 1660 |
| Roman | *Rom* | 9 | 83 | 558 | 14 | 244 | 1574 |
| Tamil | *Tam* | 3 | 150 | 873 | 36 | 240 | 451 |
| Telugu | *Tel* | 3 | 123 | 640 | 32 | 264 | 1261 |
| Thai | *Tha* | 4 | 158 | 1828 | 27 | 194 | 1856 |
| | Total: | 67 | 1396 | 8871 | 397 | 3716 | 21,974 |

Docs: number of documents; Lines: number of lines; Words: number of words

### Used Metrics

To evaluate the experiments, we utilize Cumulative Match ing Curves (CMC) [42], which measure the effectiveness of a recognition system in ranking correct matches against incorrect ones. The rank corresponds to the position at which the correct match is found within a list of potential matches. The accuracy values presented in the article's tables correspond to the rank-1 in the CMC curve.

## Experimental Results

In this section, different benchmark results are provided to get the comparative idea of different results obtained from our experiments.

### Benchmark 1: Handcrafted Feature Combination (LBP+quad-tree)

In the present benchmarking, the classifier combines two script identifiers at the score level. The first script identifier is based on LBP features and a Support Vec tor Machine, while the second relies on Quad-Tree fea tures and a Support Vector Machine. The score level combination is carried out, weighting each score at 50%. The following are the three experiments conducted in this benchmarking.

Table 5 displays the Hit Ratio of each script identifier in Benchmark 1 for the three tasks and the different training and test options.

For task 1, there are six options in the training and six options in the test, which comes to a tally of 36 different experiments shown in Table 5. Their CMC are depicted in Fig. 9.

As expected, the performance with printed text was better than that with handwritten text, probably because of the lower variability in the printed text. Also, the line-based test offered the best performance, possibly because lines contain enough information laid out in a straightforward structure. Indeed, for the printed and handwritten document cases, the classifiers trained with words work better with lines than words. This could be because line features are more stable than word features.

There is a significant decrease in the hit ratio when the training and testing images do not belong to the same case. For this reason, it was decided to train the classifier with documents, lines, and words (task 2) to build a classifier more robust to the input type: document or line or word.

In the case of task 2, according to the evaluation protocol, 12 results were obtained and are given in Table 5, while the CMC curves are shown in Fig. 9.

On average, the result of the second experiment, i.e., the procedure for training a classifier for printed and handwrit ten text, including all documents, lines, and words from the training sequence, gives a better performance than for the first experimental protocol. Similar to the first experiment, the best results were obtained when testing with lines.

*Table 5 Hit Ratio of each script identifier in Benchmark 1. The best performances for each task and training option are highlighted in bold. The results are obtained by combining two script identifiers at the score level: LBP features and a Support Vector Machine with Quad-Tree features and a Support Vector Machine.*

| Train with | Test with | | | | | |
| --- | --- | --- | --- | --- | --- | --- |
| | Handwritten | | | Printed | | |
| | Docs | Lines | Words | Docs | Lines | Words |
| **Task 1** | | | | | | |
| Handwritten docs | **79.30%** | 16.96% | 5.58% | 22.75% | 7.02% | 4.85% |
| Handwritten lines | 60.83% | **87.04%** | 54.52% | 19.66% | 25.02% | 11.11% |
| Handwritten words | 48.09% | 88.50% | **84.02%** | 37.36% | 33.04% | 26.69% |
| Printed docs | 35.03% | 37.09% | 31.57% | **90.73%** | 78.48% | 41.40% |
| Printed lines | 21.02% | 16.14% | 18.90% | 45.51% | **94.41%** | 77.17% |
| Printed words | 17.83% | 23.76% | 30.73% | 44.94% | 94.46% | **86.36%** |
| **Task 2** | | | | | | |
| Handwritten docs, lines and words | **81.21%** | **92.49%** | **83.10%** | 35.67% | 32.09% | 26.63% |
| Printed docs, lines and words | 35.35% | 34.10% | 36.97% | **88.20%** | **94.55%** | **86.55%** |
| **Task 3** | | | | | | |
| Printed and Handwritten docs, lines and words | 79.62% | 91.96% | 83.08% | 89.33% | 94.71% | 87.52% |

Moving on to task 3, the six results of the six experiments carried out, are given in Table 5, and the CMC curves are shown in Fig. 9. Similar trends are found in the results: the best result is obtained at the line level, while printed text outperforms the handwritten scenario. A confusion matrix is shown in Table 6.

The main confusions seen here are between Kannada and Telugu, Telugu and Bengali, Gujarati and Thai, and Oriya and Bengali, as shown in Fig. 10.

We have prioritized Task 3 as it yielded the best results and is considered the most valuable. Therefore, we have con ducted a detailed analysis of this specific task. Additionally, the analysis of the confusion matrices for tasks 1 and 2 led to similar conclusions.

### Benchmark 2: Handcrafted Feature (Dense Multi-Block LBP)

The second benchmark uses an SVM classifier with Dense Multi-Block LBP features. The three experiments performed in the previous benchmark were repeated in this one. All results from the second benchmark are highlighted in Table 7. Similarly to the Benchmark 1, the Table 7 presents the Hit Ratio of each script identifier in Benchmark 2 for the three tasks and the different training and test options. Regarding task 1, and similarly to the previous Benchmark 1, the performance with printed text was better than with handwritten text because of the lower intra-class variability in the printed text. Moreover, the performance at the line level was more accurate than at the document and word levels.

Besides, in the cross-document scenario, a similar pattern with Benchmark 1 can be seen. On the other hand, the best results were obtained when training with printed and tested with handwritten text. Overall, better results were achieved by Benchmark 2 versus Benchmark 1.

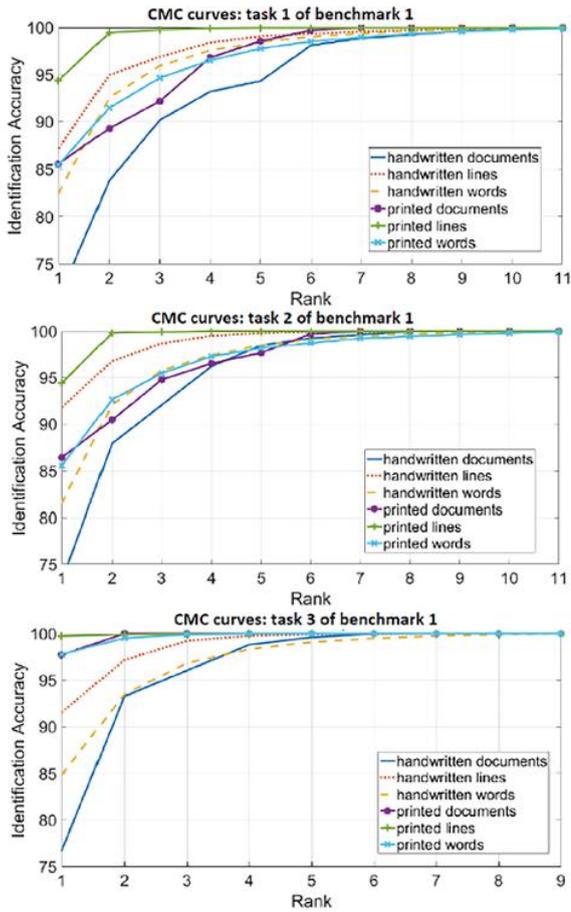

*Fig. 9 CMC curves of the three tasks of Benchmark 1. These CMC curves correspond to the results in bold in Table 5*

*Table 6 Confusion Matrix of Benchmark 1 Task 3 for Handwritten Lines, represented as a percentage of the accuracy rate*

|      | Arab  | Ban   | Guj   | Gurm  | Hind  | Jap   | Kan   | Mal   | Ori   | Rom   | Tam   | Tel   | Tha   |
|------|-------|-------|-------|-------|-------|-------|-------|-------|-------|-------|-------|-------|-------|
| Arab | 99.44 | 0.00  | 0.00  | 0.19  | 0.00  | 0.00  | 0.00  | 0.00  | 0.00  | 0.37  | 0.00  | 0.00  | 0.00  |
| Ban  | 0.49  | 89.24 | 0.00  | 2.17  | 0.07  | 0.77  | 0.00  | 0.00  | 5.24  | 1.68  | 0.07  | 0.28  | 0.00  |
| Guj  | 0.00  | 0.00  | 33.34 | 0.00  | 0.00  | 0.00  | 0.00  | 11.11 | 0.00  | 0.00  | 22.22 | 11.11 | 22.22 |
| Gurm | 0.00  | 0.00  | 0.00  | 71.43 | 28.57 | 0.00  | 0.00  | 0.00  | 0.00  | 0.00  | 0.00  | 0.00  | 0.00  |
| Hind | 0.00  | 2.04  | 0.00  | 0.00  | 95.92 | 0.00  | 2.04  | 0.00  | 0.00  | 0.00  | 0.00  | 0.00  | 0.00  |
| Jap  | 0.00  | 0.00  | 0.00  | 0.00  | 0.00  | 96.00 | 0.00  | 4.00  | 0.00  | 0.00  | 0.00  | 0.00  | 0.00  |
| Kan  | 0.00  | 0.00  | 0.00  | 0.00  | 0.00  | 0.00  | 94.13 | 0.78  | 0.00  | 0.00  | 2.74  | 2.35  | 0.00  |
| Mal  | 0.00  | 0.00  | 0.00  | 0.00  | 0.00  | 0.00  | 6.97  | 90.70 | 2.33  | 0.00  | 0.00  | 0.00  | 0.00  |
| Ori  | 0.18  | 5.52  | 0.00  | 0.46  | 0.00  | 0.18  | 1.84  | 0.09  | 89.16 | 1.10  | 0.46  | 1.01  | 0.00  |
| Rom  | 1.05  | 0.15  | 0.15  | 0.00  | 0.15  | 0.15  | 0.00  | 0.30  | 0.15  | 97.75 | 0.00  | 0.15  | 0.00  |
| Tam  | 0.00  | 0.00  | 0.00  | 0.00  | 0.00  | 0.00  | 0.00  | 3.97  | 0.00  | 0.00  | 96.03 | 0.00  | 0.00  |
| Tel  | 0.00  | 0.00  | 0.00  | 0.00  | 0.00  | 0.00  | 6.45  | 0.00  | 0.00  | 0.00  | 0.00  | 93.55 | 0.00  |
| Tha  | 0.00  | 0.00  | 0.00  | 0.00  | 0.00  | 0.00  | 0.32  | 4.76  | 0.00  | 0.00  | 0.00  | 6.98  | 87.94 |

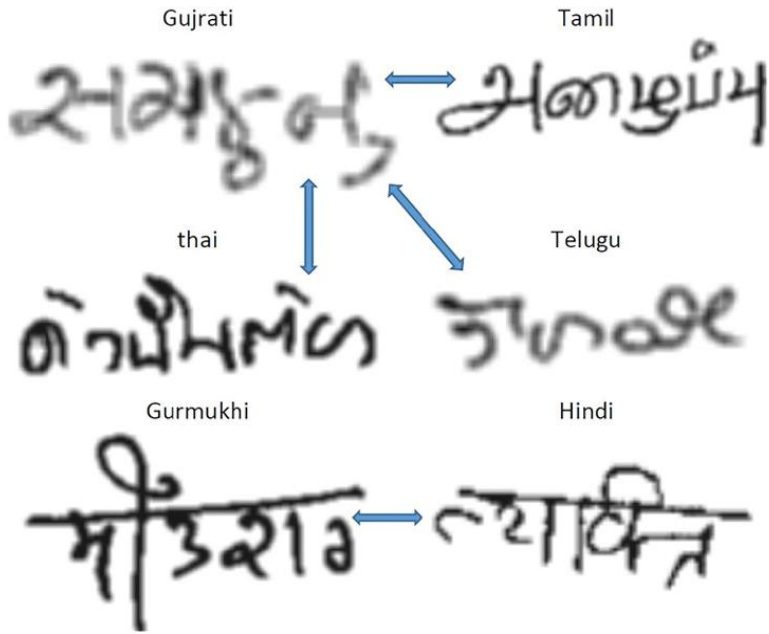

Fig. 10 Samples of the most confused scripts in Benchmark 1, Task 3. Arrows mean most common confusion

Table 7 Hit Ratio of each script identifier in Benchmark 2. The best performances for each task and training option are highlighted in bold. The results are obtained using the SVM classifier with Dense Multi Block LBP features

| Train with | Test with | | | | | |
|---|---|---|---|---|---|---|
| | Handwritten | | | Printed | | |
| | Docs | Lines | Words | Docs | Lines | Words |
| **Task 1** | | | | | | |
| Handwritten docs | **82.01%** | 20.54% | 7.46% | 26.67% | 10.04% | 7.11% |
| Handwritten lines | 65.93% | **89.78%** | 59.25% | 23.44% | 21.02% | 9.01% |
| Handwritten words | 69.92% | 89.89% | **88.01%** | 36.15% | 34.64% | 21.67% |
| Printed docs | 32.23% | 38.45% | 36.78% | **89.23%** | 80.99% | 47.83% |
| Printed Lines | 26.78% | 19.01% | 17.89% | 47.04% | **95.51%** | 79.83% |
| Printed words | 19.81% | 28.90% | 31.72% | 49.67% | 96.11% | **88.06%** |
| **Task 2** | | | | | | |
| Handwritten docs, lines and words | **83.27%** | **93.45%** | **86.51%** | 39.89% | 34.70% | 29.04% |
| Printed docs, lines and words | 37.65% | 35.16% | 39.78% | **90.23%** | **95.25%** | **89.33%** |
| **Task 3** | | | | | | |
| Printed and Handwritten docs, lines and words | 80.90% | 92.33% | 86.71% | 91.23% | 96.70% | 88.01% |

In task 2, a similar pattern of results was found to those of Benchmark 1; and the results achieved in the scenario mainly were better than those in Benchmark 1. Similar to the two previous sets of experiments, in the third task of Benchmark 2, a similar pattern was found, with better accuracy than in Benchmark 1.

***Benchmark 3: Deep Neural Networks***

The third benchmark was carried out with the above-mentioned DNN architectures. For a fair comparison, the experimental protocol proposed for the previous benchmarks was repeated. All the results obtained for this third benchmark are included in Table 9. The ResNet architecture clearly out performs the VGG architecture with a performance improvement of 2-4% and 10% for printed and handwritten samples, respectively. The rest of the analysis will be focused on the performance of the ResNet model.

Task 1 with Deep Neural Networks showed a very competitive performance for printed samples. The results obtained outperformed the first benchmarks for printed data. As in previous experiments, lines showed the best performance, followed by words and documents. When large databases are available, deep representations are capable of achieving almost 99% accuracy for printed patterns.

The performance obtained for handwritten samples was similar to the first experiments with the Benchmarks 1 and 2. The gap between the performance obtained for printed samples and handwritten samples is caused by the large intra-class variability of the writers. The Deep Neural Net works are unable to reach a good generalization because of this larger variability. There is room for improvement and training deep representations capable of dealing with writer variability is a key challenge in this area. The MDIW-13 provides an extensive multi-lingual database to train and evaluate these models. For the second task, Deep Neural Networks achieved the best performances with printed samples. Once again, the performance obtained for handwritten samples was poor in comparison with the other two benchmarks.

*Table 8 Comparison of Hit Ratio for VGG and ResNet Architectures in Benchmark 3. Accuracies are obtained averaging the results marked with bold font in Table 9*

|  | Task 1 | | Task 2 | | Task 3 | |
|---|---|---|---|---|---|---|
|  | Handwritten | Printed | Handwritten | Printed | Handwritten | Printed |
| VGG | **77.69%** | **94.06%** | **79.36%** | **94.82%** | **27.64%** | **94.74%** |
| ResNet | **85.35%** | **96.51%** | **89.28%** | **98.41%** | **34.50%** | **98.45%** |

The more significant number of samples used here produced a slight improvement for printed samples.

In task 3, unlike the above benchmarks, the results in the printed case were not improved and produced a clear drop in performance in the handwritten case. These results suggest that handwritten and printed models should be trained separately for Deep Neural Networks. As commented before, writer variability is not well modelled by the DNN. Therefore, it is clear that the training strategy depends on the classifier and the features in comparing the three benchmarks.

Finally, Table 8 compares the performance achieved by the two Deep Neural Network architectures evaluated. The Hit Ratio for each task and type of sample was obtained by averaging the Hit Ratios obtained when the training and test samples belong to the same class (e.g., handwritten documents). The results averaged in Table 8 correspond to the average of the results highlighted with bold font in Table 9. Similarly to the Benchmark 1 and 2, the Table 9 displays the Hit Ratio of each script identifier in Benchmark 3 for the three tasks and the different training and test options. The results show the superior performance of the ResNet architecture with performance improvement of around 10% for handwritten experiments and 2-4% for experiments with printed samples. These results encourage us to find new Deep Neural Network architectures capable of modelling the variability in handwritten classification

.

*Table 9 Hit Ratio of each script identifier in Benchmark 3 - ResNet Model. The best performances for each task and training option are highlighted in bold*

| Train with | Test with | | | | | |
|---|---|---|---|---|---|---|
| | Handwritten | | | Printed | | |
| | Docs | Lines | Words | Docs | Lines | Words |
| **Task 1** | | | | | | |
| Handwritten docs | **78.43%** | 47.54% | 39.49% | 27.14% | 28.96% | 27.87% |
| Handwritten lines | 47.41% | **89.92%** | 71.68% | 33.02% | 30.09% | 29.35 |
| Handwritten words | 47.38% | 87.91% | **87.72%** | 42.17% | 48.63% | 43.75% |
| Printed docs | 18.19% | 29.01% | 25.55% | **93.55%** | 95.64% | 85.16% |
| Printed Lines | 19.88% | 30.82% | 28.08% | 91.28% | **99.53%** | 95.67% |
| Printed words | 18.02% | 29.67% | 30.56% | 90.28% | 96.81% | **96.46%** |
| **Task 2** | | | | | | |
| Handwritten docs, lines and words | **86.24%** | **92.48%** | **89.14%** | 52.58% | 52.02% | 45.61% |
| Printed docs, lines and words | 21.49% | 36.87% | 30.93% | **96.84%** | **99.82%** | **98.57%** |
| **Task 3** | | | | | | |
| Printed and Handwritten docs, lines and words | 29.26% | **40.04%** | **34.20%** | 96.48% | 99.78% | 99.09% |

## Discussion

Globally speaking, this paper aimed to introduce a new multilingual and multi-script database that allows the development of new algorithms, applications, and a simple and easy-to-go benchmark to facilitate the comparison [43, 44].

The benchmarking reveals some new possibilities of using the database. For instance, the division in documents, lines, and words enables the training of a script model with a level, for instance, lines, and testing at other levels, for instance, words. The results obtained show that the technology requires improvements due to the lack of generalization of the identifiers when moving the test from one level, e.g. words, to another, e.g. documents.

Furthermore, the benchmarking highlight an interesting direction when training the model with images from all the levels and testing with images of different levels. Furthermore, the model with the best identification rates at the three levels in the three conducted experiments is the model trained with documents and lines plus words. It suggests that general identifiers at the three levels are possible and how to train them in practical applications. Even if the lines are obtained from the documents and the words from the lines or an artificial line or document are build up from words or lines.

Instead, a global model for printed and handwritten is still far from reasonable results, mainly in the case of deep learn ing [45], at least with the well-established classifiers used in this work.

Regarding the benchmark, the idea of a simple and easy to-go benchmark to facilitate the comparison has its limitations. To this aim, training and testing set as functions to calculate the parameters and implement the classifiers have been defined. It leads to repeatability research since the used methods are easy to find in scientific free software packages. From now onwards, developing new state-of-the-art script identifiers and improving database partition is a task done by the researcher enticed by this new public database. Further works should be done to explore novel data-driven learning frameworks. This research line includes novel architectures as well as new learning frameworks, including synthetic data to improve the generalization capacity of the models (e.g., Generative Adversarial Networks or Variational Autoencoders). Obviously, this database can also be enlarged with new scripts and more samples from the scripts to make it more appealing.

**Conclusion**

A new multi-lingual and multi-script dataset (MDIW-13) for script identification, including printed and handwritten documents for free distribution, is introduced in this paper. The handwritten part was collected from letters or notes developed by volunteers living in the native zones where scripts were gathered. These volunteers scanned their documents and sent them in by e-mail. The printed samples were obtained from local newspaper and magazines and contain different fonts and sizes and cursive and bold text. The printed documents were scanned at 300 dpi.

Because the database targeted script identification tasks, the document background was converted to white, and the text ink equalized to avoid watermarks due to the local paper or ink textures, which could bias the results of a script identifier. This procedure was manually monitored.

MDIW-13 allows experiments with script identification at different levels (e.g., document, lines, and words). To this aim, the lines of each document were extracted from the documents and the words from the lines.

Three benchmarks were conducted. The first one relies on local descriptors such as LBP and Quad-Tree histograms with an SVM. The second one is based on Dense Multi Block LBPs, and produces excellent results due to their multi-scale and denser spatial description. The third bench mark is based on two Deep Neural Network architectures. The benchmark includes results at the document, line, and word levels, in addition to providing results at the handwrit ten and printed levels. Finally, they give results of a script identifier independent of the handwritten or printed text level at play.

It is expected that this new multi-lingual database will elicit new script identifiers, open the door to developing new problems like challenges in writer dependent or independent script identification challenges with the handwritten part of the dataset, artistic multi-character script identification [46], or advanced algorithms for segmenting handwritten and printed-based images and allow new insights into script identification. The different scenarios in the present study, including handwritten and printed samples, reveal numerous challenges. The results reported for the three bench marks could serve as a baseline for further research in script identification.

Future work with this database might include but is not limited to: i) the analysis of hybrid models based on both statistical approaches and deep features; ii) the use of novel architectures (e.g., CNN-LSTM, VAE) to incorporate context in the learning process of visual features; iii) the application of domain adaptation techniques to employ pre trained models that take advantages of embedding spaces learned from similar domains (e.g., text classification).

**Additional Results**

Tables 10, 11, 12, 13, 14, and 15, which contain additional results, are included in the appendix of this article.

*Table 10 Database statistics at document level*

| Script abrev | Handwritten | | | | | | Printed | | | | | |
|---|---|---|---|---|---|---|---|---|---|---|---|---|
| | **NumDoc** | **NumPix** | **AvgNumPix** | **StdNumPix** | **MinNumPix** | **MaxNumPix** | **NumDoc** | **NumPix** | **AvgNumPix** | **StdNumPix** | **MinNumPix** | **MaxNumPix** |
| Arab | 48 | 16364808 | 340934 | 140942 | 24741 | 692100 | 51 | 7673662 | 150464 | 79869 | 53637 | 365978 |
| Ban | 67 | 57354267 | 856034 | 258275 | 269796 | 1408974 | 51 | 3527860 | 69174 | 32182 | 15288 | 147683 |
| Guj | 3 | 1112702 | 370901 | 26643 | 346901 | 399570 | 32 | 2798190 | 87443 | 65878 | 23870 | 390789 |
| Gurm | 6 | 3190761 | 531794 | 304299 | 21977 | 818942 | 115 | 9653569 | 83944 | 76390 | 21154 | 476005 |
| Hind | 21 | 4078695 | 194224 | 180115 | 9463 | 526492 | 47 | 2887991 | 61447 | 44688 | 13356 | 214140 |
| Jap | 20 | 9366326 | 468316 | 136,807 | 292364 | 724561 | 80 | 2367889 | 29599 | 13560 | 9069 | 71639 |
| Kan | 15 | 5676355 | 378424 | 284863 | 160551 | 864733 | 53 | 3170929 | 59829 | 33427 | 14487 | 192797 |
| Mal | 12 | 1952046 | 162671 | 157549 | 25357 | 457572 | 70 | 4273222 | 61046 | 35702 | 12074 | 198668 |
| Ori | 50 | 45562745 | 911255 | 260843 | 319114 | 1416357 | 42 | 3338941 | 79499 | 94056 | 10581 | 483419 |
| Rom | 90 | 22377054 | 248634 | 82037 | 59977 | 490772 | 56 | 11527272 | 205844 | 106145 | 56905 | 478682 |
| Tam | 14 | 9345399 | 667529 | 130809 | 435632 | 870280 | 46 | 1075869 | 23388 | 18041 | 3090 | 84282 |
| Tel | 10 | 9165463 | 916546 | 633586 | 236418 | 1965304 | 49 | 3181893 | 64937 | 45549 | 14928 | 232850 |
| Tha | 26 | 14004980 | 538653 | 102752 | 342289 | 685656 | 61 | 4402236 | 72168 | 61943 | 24270 | 445787 |
| Total | 382 | 199551601 | 506609 | 207655 | 9463 | 1965304 | 753 | 59879523 | 80675 | 54418 | 3090 | 483419 |

Numdoc: Number of documents. NumPix: Total number of pixels in the documents. AvgNumPix: Averaged number of pixels per document. StdNumPix: Standard deviation of the number of pixels per document. MinNumPix: Minimum number of pixels in a document. MaxNumPix: Maximum number of pixels in a document

*Table 11 Database statistics at line level*

| Script abrev | Handwritten | | | | | | Printed | | | | | |
|---|---|---|---|---|---|---|---|---|---|---|---|---|
| | **NumLines** | **NumPix** | **AvgNumPix** | **StdNumPix** | **MinNumPix** | **MaxNumPix** | **NumLines** | **NumPix** | **AvgNumPix** | **StdNumPix** | **MinNumPix** | **MaxNumPix** |
| Arab | 621 | 13276894 | 21380 | 9549 | 2887 | 98922 | 1082 | 7336656 | 6781 | 1854 | 1766 | 11895 |
| Ban | 1486 | 51832253 | 34880 | 8973 | 5590 | 63786 | 466 | 3549672 | 7617 | 5202 | 2336 | 70654 |
| Guj | 41 | 302810 | 7386 | 2777 | 4031 | 13424 | 384 | 4464622 | 11627 | 20822 | 1916 | 191531 |
| Gurm | 111 | 1368759 | 12331 | 4631 | 1673 | 19407 | 1062 | 9461641 | 8909 | 9475 | 2453 | 34604 |
| Hind | 230 | 1858972 | 8082 | 4542 | 1106 | 20159 | 397 | 3510591 | 8843 | 7755 | 2344 | 80344 |
| Jap | 121 | 968815 | 8007 | 5076 | 1735 | 29604 | 559 | 2215839 | 3964 | 991 | 1311 | 10639 |
| Kan | 377 | 5105947 | 13544 | 5223 | 2834 | 29903 | 582 | 3933958 | 6759 | 10509 | 2012 | 190604 |
| Mal | 211 | 2398611 | 11368 | 5075 | 1849 | 22886 | 706 | 3967304 | 5619 | 1895 | 1181 | 11749 |
| Ori | 1136 | 41062961 | 36147 | 8607 | 6875 | 66079 | 548 | 3163802 | 5773 | 2921 | 2447 | 43399 |
| Rom | 750 | 18868245 | 25158 | 6906 | 7916 | 46564 | 961 | 10348478 | 10768 | 3423 | 2259 | 15838 |
| Tam | 276 | 3553407 | 12875 | 3947 | 2720 | 24200 | 301 | 2193811 | 7288 | 2958 | 977 | 20873 |
| Tel | 154 | 2283067 | 14825 | 5947 | 2781 | 25639 | 483 | 3766077 | 7797 | 9809 | 2065 | 134336 |
| Tha | 473 | 6658573 | 14077 | 4989 | 2886 | 25598 | 461 | 4288718 | 9303 | 2487 | 3608 | 18001 |
| Total | 5987 | 149539314 | 16928 | 5865 | 1106 | 98922 | 7992 | 62201169 | 7773 | 6162 | 977 | 191531 |

NumLines: Number of lines. NumPix: Total number of pixels in the lines. AvgNumPix: Averaged number of pixels per line. StdNumPix: Standard deviation of the number of pixels per word. MinNumPix: Minimum number of pixels in a word. MaxNumPix: Maximum number of pixels in a word

*Table 12 Database statistics at word level*

| Script abrev | Handwritten | | | | | | Printed | | | | | |
|---|---|---|---|---|---|---|---|---|---|---|---|---|
| | **NumWords** | **NumPix** | **AvgNumPix** | **StdNumPix** | **MinNumPix** | **MaxNumPix** | **NumWords** | **NumPix** | **AvgNumPix** | **StdNumPix** | **MinNumPix** | **MaxNumPix** |
| Arab | 3940 | 13464275 | 3417 | 742 | 2233 | 7363 | 6202 | 6356966 | 1025 | 310 | 373 | 2833 |
| Ban | 9320 | 44844887 | 4812 | 2061 | 1239 | 21767 | 2557 | 3364773 | 1316 | 1273 | 389 | 42379 |
| Guj | 181 | 259737 | 1435 | 622 | 349 | 3321 | 2211 | 4151893 | 1878 | 3607 | 379 | 56110 |
| Gurm | 700 | 1346638 | 1924 | 884 | 529 | 6184 | 9104 | 8123326 | 892 | 556 | 155 | 4233 |
| Hind | 1457 | 1317882 | 905 | 579 | 204 | 5087 | 2782 | 3157311 | 1135 | 1556 | 333 | 33827 |
| Jap | 441 | 1084912 | 2460 | 1721 | 536 | 14947 | 1814 | 1770894 | 976 | 326 | 342 | 2545 |
| Kan | 1995 | 4809571 | 2411 | 1201 | 501 | 7576 | 2157 | 3550616 | 1646 | 2565 | 442 | 61944 |
| Mal | 719 | 2410708 | 3353 | 1808 | 572 | 11309 | 4320 | 3425423 | 793 | 283 | 333 | 2451 |
| Ori | 7847 | 38539556 | 4911 | 2210 | 1480 | 20991 | 2309 | 2988126 | 1294 | 791 | 395 | 13818 |
| Rom | 4308 | 16591698 | 3851 | 2000 | 758 | 16830 | 7627 | 9692418 | 1271 | 690 | 338 | 4687 |
| Tam | 1430 | 3443374 | 2408 | 1115 | 619 | 8534 | 2118 | 7201309 | 3400 | 3099 | 353 | 33514 |
| Tel | 801 | 2036423 | 2542 | 1282 | 448 | 8404 | 2126 | 3654671 | 1719 | 2269 | 371 | 42535 |
| Tha | 4472 | 5914351 | 1323 | 730 | 348 | 6846 | 3717 | 3948106 | 1062 | 470 | 322 | 3345 |
| Total | 37611 | 136064012 | 2750 | 1304 | 204 | 21767 | 49044 | 61385832 | 1416 | 1369 | 155 | 61944 |

NumWords: Number of words. NumPix: Total number of pixels in the words. AvgNumPix: Averaged number of pixels per word. StdNumPix: Standard deviation of the number of pixels per word. MinNumPix: Minimum number of pixels in a word. MaxNumPix: Maximum number of pixels in a word

*Table 13 Training statistics at document level*

| Script abrev | Handwritten | | | | | | Printed | | | | | |
|---|---|---|---|---|---|---|---|---|---|---|---|---|
| | **NumDoc** | **NumPix** | **TrainDoc** | **%TrainDoc** | **NumTrainPix** | **%NumTrainPix** | **NumDoc** | **NumPix** | **TrainDoc** | **%TrainDoc** | **NumTrainPix** | **%NumTrainPix** |
| Arab | 48 | 16364808 | 5 | 10.42% | 2420784 | 14.79% | 51 | 7673662 | 14 | 27.45% | 2117246 | 27.59% |
| Ban | 67 | 57354267 | 3 | 4.48% | 3120881 | 5.44% | 51 | 3527860 | 27 | 52.94% | 2018089 | 57.20% |
| Guj | 3 | 1112702 | 2 | 66.67% | 765801 | 68.82% | 32 | 2798190 | 22 | 68.75% | 2074532 | 74.14% |
| Gurm | 6 | 3190761 | 4 | 66.67% | 2471223 | 77.45% | 115 | 9653569 | 39 | 33.91% | 2044016 | 21.17% |
| Hind | 21 | 4078695 | 15 | 71.43% | 2010846 | 49.30% | 47 | 2887991 | 33 | 70.21% | 2107547 | 72.98% |
| Jap | 20 | 9366326 | 4 | 20.00% | 2101719 | 22.44% | 80 | 2367889 | 64 | 80.00% | 1906104 | 80.50% |
| Kan | 15 | 5676355 | 3 | 20.00% | 2169454 | 38.22% | 53 | 3170929 | 38 | 71.70% | 2025503 | 63.88% |
| Mal | 12 | 1952046 | 9 | 75.00% | 1389695 | 71.19% | 70 | 4273222 | 26 | 37.14% | 2010067 | 47.04% |
| Ori | 50 | 45562745 | 3 | 6.00% | 2559961 | 5.62% | 42 | 3338941 | 25 | 59.52% | 2279637 | 68.27% |
| Rom | 90 | 22377054 | 9 | 10.00% | 2060441 | 9.21% | 56 | 11527272 | 14 | 25.00% | 2241415 | 19.44% |
| Tam | 14 | 9345399 | 3 | 21.43% | 2277994 | 24.38% | 46 | 1075869 | 36 | 78.26% | 910447 | 84.62% |
| Tel | 10 | 9165463 | 3 | 30.00% | 2589472 | 28.25% | 49 | 3181893 | 32 | 65.31% | 2041792 | 64.17% |
| Tha | 26 | 14004980 | 4 | 15.38% | 2288212 | 16.34% | 61 | 4402236 | 27 | 44.26% | 2204803 | 50.08% |
| Total | 382 | 199551601 | 67 | 17.54% | 28226483 | 14.14% | 753 | 59879523 | 397 | 52.72% | 25981198 | 43.39% |

Numdoc: Number of documents. NumPix: Total number of pixels in the documents. TrainDoc: Number of documents in the training set

%TrainDoc: Percentage of documents in the training. NumTrainPix: Number of pixels in the training. %NumTrainPix: Percentage of pixels in the training

*Table 14 Training statistics at line level*

| Script abrev | Handwritten | | | | | | Printed | | | | | |
|---|---|---|---|---|---|---|---|---|---|---|---|---|
| | NumLines | NumPix | TrainLines | %TrainLines | NumTrainPix | %NumTrainPix | NumLines | NumPix | TrainLines | %TrainLines | NumTrainPix | %NumTrainPix |
| Arab | 621 | 13276894 | 88 | 14.17% | 2010497 | 15.14% | 1082 | 7336656 | 256 | 23.66% | 2005281 | 27.33% |
| Ban | 1486 | 51832253 | 55 | 3.7% | 2026720 | 3.91% | 466 | 3549672 | 234 | 50.21% | 2000904 | 56.37% |
| Guj | 41 | 302810 | 32 | 78.05% | 260126 | 85.9% | 384 | 4464622 | 190 | 49.48% | 2002409 | 44.85% |
| Gurm | 111 | 1368759 | 88 | 79.28% | 1121767 | 81.96% | 1062 | 9461641 | 468 | 44.07% | 2004257 | 21.18% |
| Hind | 230 | 1858972 | 184 | 80,00% | 1489073 | 80.1% | 397 | 3510591 | 215 | 54.16% | 2006034 | 57.14% |
| Jap | 121 | 968815 | 96 | 79.34% | 821819 | 84.83% | 559 | 2215839 | 447 | 79.96% | 1796251 | 81.06% |
| Kan | 377 | 5105947 | 122 | 32.36% | 2002821 | 39.23% | 582 | 3933958 | 302 | 51.89% | 2003321 | 50.92% |
| Mal | 211 | 2398611 | 168 | 79.62% | 1888048 | 78.71% | 706 | 3967304 | 314 | 44.48% | 2003869 | 50.51% |
| Ori | 1136 | 41062961 | 49 | 4.31% | 2030372 | 4.94% | 548 | 3163802 | 348 | 63.5% | 2005057 | 63.37% |
| Rom | 750 | 18868245 | 83 | 11.07% | 2001798 | 10.61% | 961 | 10348478 | 244 | 25.39% | 2009359 | 19.42% |
| Tam | 276 | 3553407 | 150 | 54.35% | 2004029 | 56.4% | 301 | 2193811 | 240 | 79.73% | 1697694 | 77.39% |
| Tel | 154 | 2283067 | 123 | 79.87% | 1700793 | 74.5% | 483 | 3766077 | 264 | 54.66% | 2002136 | 53.16% |
| Tha | 473 | 6658573 | 158 | 33.4% | 2018174 | 30.31% | 461 | 4288718 | 194 | 42.08% | 2003650 | 46.72% |
| Total | 5987 | 149539314 | 1396 | 23.32% | 21376037 | 14.29% | 7992 | 62201169 | 3716 | 46.5% | 25540222 | 41.06% |

NumLines: Number of lines. NumPix: Total number of pixels in the lines. TrainLines: Number of lines in the training set

%TrainLines: Percentage of lines in the training. NumTrainPix: Number of pixels in the training. %NumTrainPix: Percentage of pixels in the training

*Table 15 Training statistics at word level*

| Script abrev | Handwritten | | | | | | Printed | | | | | |
|---|---|---|---|---|---|---|---|---|---|---|---|---|
| | NumWords | NumPix | TrainWords | %Train-Words | NumTrain-Pix | %NumTrain-Pix | NumWords | NumPix | TrainWords | %Train-Words | NumTrain-Pix | %NumTrain-Pix |
| Arab | 3940 | 13464275 | 570 | 14.47% | 2000893 | 14.86% | 6202 | 6356966 | 1996 | 32.18% | 2001192 | 31.48% |
| Ban | 9320 | 44844887 | 401 | 4.3% | 2003873 | 4.47% | 2557 | 3364773 | 1608 | 62.89% | 2000563 | 59.46% |
| Guj | 181 | 259737 | 144 | 79.56% | 224064 | 86.27% | 2211 | 4151893 | 1229 | 55.59% | 2001494 | 48.21% |
| Gurm | 700 | 1346638 | 560 | 80% | 1103880 | 81.97% | 9104 | 8123326 | 3629 | 39.86% | 2000388 | 24.63% |
| Hind | 1457 | 1317882 | 1165 | 79.96% | 996000 | 75.58% | 2782 | 3157311 | 1706 | 61.32% | 2000318 | 63.36% |
| Jap | 441 | 1084912 | 352 | 79.82% | 928975 | 85.63% | 1814 | 1770894 | 1451 | 79.99% | 1420649 | 80.22% |
| Kan | 1995 | 4809571 | 872 | 43.71% | 2000559 | 41.6% | 2157 | 3550616 | 1183 | 54.84% | 2000213 | 56.33% |
| Mal | 719 | 2410708 | 575 | 79.97% | 1870088 | 77.57% | 4320 | 3425423 | 2370 | 54.86% | 2000079 | 58.39% |
| Ori | 7847 | 38539556 | 333 | 4.24% | 2001919 | 5.19% | 2309 | 2988126 | 1660 | 71.89% | 2000920 | 66.96% |
| Rom | 4308 | 16591698 | 558 | 12.95% | 2002681 | 12.07% | 7627 | 9692418 | 1574 | 20.64% | 2000076 | 20.64% |
| Tam | 1430 | 3443374 | 873 | 61.05% | 2001245 | 58.12% | 2118 | 7201309 | 451 | 21.29% | 2011290 | 27.93% |
| Tel | 801 | 2036423 | 640 | 79.9% | 1511017 | 74.2% | 2126 | 3654671 | 1261 | 59.31% | 2000984 | 54.75% |
| Tha | 4472 | 5914351 | 1828 | 40.88% | 2000091 | 33.82% | 3717 | 3948106 | 1856 | 49.93% | 2000011 | 50.66% |
| Total | 37611 | 136064012 | 8871 | 23.59% | 20645285 | 15.17% | 49044 | 61385832 | 21974 | 44.8% | 25438177 | 41.44% |

NumWords: Number of words. NumPix: Total number of pixels in the words. TrainWords: Number of words in the training set

%TrainWords: Percentage of words in the training. NumTrainPix: Number of pixels in the training. %NumTrainPix: Percentage of pixels in the training